# Machine learning and digital pragmatics: Which word category influences emoji use most?

Mohammed Q. Shormani[1], Lamia Alhassoun[2] & Yehia A. AlSohbani[3]
[1]Ibb University, Ibb, Yemen
[2&3]Arab Open University, Riyadh, Saudi Arabia
[1]shormani@ibbuniv.edu.ye/https://orcid.org/0000-0002-0138-4793
(April 2026)

## Abstract

This study examines the performance of the state-of-the-art MARBERT model in identifying the lexical/pragmatic category associated with emoji use on X within a digital pragmatics approach (DPA). A net corpus of 15856 Colloquial Arabic (CA) posts containing emojis was collected from X using Python. The texts were tokenized and normalized into 4 lexical categories, namely noun_norm, verb_norm, adj_norm, and adverb_norm, and 2 pragmatic/structural categories, question_norm and exclamation_norm. MARBERT was finetuned and optimized to identify which category scores standard metrics more, hence associated with emoji use, while binary logistic regression was used to examine which category is statistically associated with emoji occurrence. Findings unveil that nouns dominate the corpus in normalized frequency (M = 0.675, SD = 0.161), followed by verbs (M = 0.083, SD = 0.100). However, verbs have the strongest influence of emoji use indicated by verb density (β = 0.821, *p* = .001, 95% CI [0.332, 1.309]). Findings further show that MARBERT performs comparatively well for verbs (precision = .74, recall = .99, F1 = .85) and nouns (precision = .88, recall = .44, F1 = .59), and substantially weaker for the less frequent categories. The study concludes that in digital pragmatics of CA on X, emoji use association with lexical/pragmatic category can be explained by a hybrid approach of computational, statistical, and pragmatic methods, reflecting the interaction among machine learning, linguistic/lexical features, contextual representation, and digital communication.

## 1. Introduction

Machine learning (ML) is a subfield of artificial intelligence (AI). It concerns the development of computational models that learn patterns from data and make predictions or decisions without being explicitly programmed (see e.g. Ghahramani 2015; Shormani 2025). Instead of relying on fixed rules, ML algorithms infer relationships within data through training processes, enabling them to generalize to unseen instances. Additionally, digital discourse/communication (DC) has undergone a significant transformation with the rise of multimodal expression, where written language is increasingly complemented by visual symbols (see e.g. Li and Yang 2018; Kariryaa et al. 2022; Bou-Franch and Blitvich 2018; Yus 2011, 2025). Emojis are perhaps the most notable visual symbols employed in DC. Emojis could be referred to as pictogram, logogram words, drawn from the Japanese *e* for 'picture' and *moji* for 'word' (see also Scheffler et al. 2022). These "symbolic words" have become a pervasive feature of online discourse, functioning not merely as decorative elements but as meaningful semiotic resources that contribute to the interpretation of textual messages. This shift towards multimodal interaction has catalyzed the evolution of digital pragmatics. Broadly defined, digital pragmatics examines how meaning is conveyed and interpreted through visual cues, such as emojis, GIFs, and stickers, in context (Danesi 2016; Evans 2017). It focuses on how non-linguistic signs function as communicative resources that shape pragmatic interpretation (Shormani and Alenezi 2026), setting it apart from purely textual pragmatics, which focuses on meaning construction strictly through linguistic elements (cf. Bou-Franch & Blitvich 2018). Digital pragmatics emphasizes that online communication relies heavily on non-verbal surrogates to compensate for the absence of face-to-face cues like facial expressions, prosody, and physical presence (Shormani and Alenezi 2026). In place of physical cues, users employ a repertoire of semiotic resources, including expressive typography, spelling variations, images, emoticons, and emojis (Kress and van Leeuwen, 2006). Rather than serving as mere ornamental add-ons, these elements actively shape pragmatic force by indicating stance, modulating speech acts (softening or intensifying), signaling group identity, and framing utterance interpretation (Yus, 2011, 2025). As such, emojis can be conceptualized as visual pragmatic markers that operate alongside textual elements to enrich or modify meaning. Their role becomes particularly salient in informal, fast-paced digital environments where brevity and expressiveness are highly valued (see e.g. Ullah et al. 2025; Yus 2025).

Social media platforms such as X, Facebook, Instagram, and WhatsApp provide technological environments in which users produce and negotiate digital discourse. Their communicative affordances shape how linguistic, pragmatic, and multimodal resources are used in online interaction. X perhaps requires resorting to visual signs, viz., emojis, as it allows limited number of words in each post, specifically for unsubscribed/free users (Shormani and Alenezi 2026). However, these DC outlets are an essential part of our daily life across the world, and Arabs are no exception. Given that Colloquial Arabic (CA) is everyone's capability, every X user utilizes it on this platform. In this study, we conceptualize CA as any Arabic dialect acquired at home, i.e. other than (Modern) Standard Arabic (SA).  the variety targeted by the present corpus and as a form of Arabic frequently used in informal digital communication Analyzing any communication phenomenon within this sphere requires an approach that could interpret the aspects of this phenomenon, an approach based on digital pragmatics (DP) (cf. Bou-Franch and Blitvich 2018; Bou-Franch 2020; Shormani and Alenezi 2026).

In digital discourse, visual elements like emojis become particularly significant due to the absence of many non-verbal cues present in face-to-face interaction, or textual pragmatics (cf. Grice 1975; Searle 1969; Levinson 1983). It follows then that writers rely more heavily on linguistic strategies such as word choice, repetition, and structural emphasis to convey tone, emotion, and interpersonal meaning (Ullah et al. 2025; Wang et al. 2024). In this sense, textual pragmatics provides the foundation for understanding how purely linguistic elements function in communication, specifically when interacting with additional semiotic resources such as emojis (Danesi 2016). Accordingly, this study conceptualizes digital pragmatics on X as an interaction between textual pragmatics and digital pragmatics, where lexical choices and emojis jointly contribute to meaning construction. It specifically investigates whether word category influences emoji use in a corpus of CA posts. By examining the distributional relationship between lexical categories and emoji occurrence, the study aims to provide new insights into the structured nature of multimodal communication in Arabic digital contexts.

In this study, we aim to connect both concepts in relation to digital discourse, specifically on X given that textual pragmatics accounts for how meaning is encoded through linguistic choices, viz., word categories such as nouns, verbs, and adjectives, which carry varying degrees of semantic and affective content. However, on X, these cues are often pragmatically underspecified, unconsolidated, given a room for vague interpretations due to the absence of tone, facial expression, or prosody. DP resorts to emoji use as visual markers that reinforce, clarify, or sometimes alter the meaning conveyed by text.

This study integrates linguistic analysis with transformer-based computational modeling to investigate emoji occurrence in Arabic digital discourse. The linguistic component examines whether specific lexical and structural features of digital discourse are statistically associated with the occurrence of emojis, while the computational component uses MARBERT to model emoji occurrence from the same textual data. Thus, the 2 components address the same empirical phenomenon from complementary perspectives: the statistical analysis identifies and quantifies linguistic associations with emoji occurrence, whereas MARBERT evaluates the extent to which patterns in Arabic digital text can be computationally learned for emoji-occurrence prediction. This integration allows the study to connect linguistic interpretation with computational modeling rather than treating them as independent analytical tasks.

1. What is the relationship between lexical and structural features of ADD and emoji occurrence?
2. Which linguistic/pragmatic features are significantly associated with the occurrence of emojis use statistically?
3. Can MARBERT, an Arabic-based Transformer-based model, computationally predict emoji occurrence from the same digital discourse data and what is its predictive performance using standard classification metrics?

Thus, this article is structured as follows. Section 2 presents theoretical foundations and previous studies. Section 3 outlines the study methods. Section 4 presents the study results. Section 5 discusses these results, and Section 6 concludes the article presenting some limitations.

## 2. Theoretical foundations and previous studies

### 2.1. Machine learning

Machine learning is a subfield of artificial intelligence concerned with the development of computational models that learn patterns from data and make predictions or decisions without being explicitly programmed (see e.g. Murphy 2012; Ghahramani 2015). Instead of relying on fixed rules, machine learning algorithms infer relationships within data through training processes, enabling them to generalize to unseen instances. Classical machine learning methods include algorithms such as logistic regression, decision trees, and support vector machines, which typically rely on structured input features designed by researchers (see e.g. Mitchell 1997). These approaches have been widely applied across domains such as natural language processing, computer vision, and social media analysis due to their interpretability and efficiency (Mitchell 1997; Murphy 2012).

In recent years, ML has evolved to incorporate more advanced approaches, particularly DL, which uses multi-layered neural networks to automatically learn representations from raw data (see e.g. Taye 2023). While DL models such as transformer architectures have demonstrated strong performance in language-related tasks, traditional machine learning methods remain essential for interpretability and hypothesis testing. In linguistic research, specifically in digital discourse analysis, combining machine learning techniques with statistical models allows researchers to both predict patterns and explain underlying relationships. This hybrid approach enhances analytical rigor by integrating predictive accuracy with theoretical interpretability (Goodfellow et al. 2016; Bishop 2006).

### 2.2. Digital discourse and pragmatics

Digital discourse refers to communication occurring across digital platforms, characterized by interactivity, multimodality, and rapid exchange (Shormani and Alenezi 2026). Digital discourse emerges as a result of the growth of "computer-mediated communication" as a distinct mode of discourse with its own structural and pragmatic features (see e.g. Jones et al. 2015). These include abbreviated language, hybrid forms, and the integration of non-verbal elements such as emojis (see e.g. Herring 2004). It encompasses how people create meaning, construct identity, and manage power dynamics online. Common examples include social media conversations, instant messaging, vlogging, and comments (KhosraviNik 2014).

The rise of social media platforms and internet has further transformed digital discourse into a highly dynamic and participatory environment. Online communication fosters linguistic innovation, including the blending of spoken and written language features (Crystal 2011). This hybridity is particularly evident in informal contexts where users prioritize expressiveness and immediacy. Digital discourse is also shaped by platform-specific affordances such as character limits and interface design (see e.g. Bucher and Helmond 2018). These constraints encourage users to adopt efficient communicative strategies, often combining text with visual elements. It then follows that meaning in digital discourse is increasingly multimodal, requiring analytical frameworks that account for both textual and visual components. One of the communicative strategies is the use of emojis (Aijmer 2013; An et al. 2018).

Perhaps digital discourse (and social media platforms) lead to the evolution of digital pragmatics. The term *digital pragmatics* could refer to what visual elements and cues like emojis, GIF, stickers express (see e.g. Danesi 2016; Evans 2017). It can be defined as the study of how meaning is conveyed and interpreted through visual elements in context, focusing on how non-linguistic signs such as images, symbols, and emojis function as communicative resources that shape pragmatic meaning (Shormani and Alenezi 2026). This contrasts it with *Textual pragmatics.* The latter can be simply defined as the study of how meaning is constructed and interpreted through linguistic elements in context (cf. Bou-Franch and Blitvich 2018). Digital pragmatics emphasizes that communication in online environments depends heavily on nonverbal cues that compensate for the absence of face-to-face elements such as facial expressions, gestures, and physical presence (Shormani and Alenezi 2026). In place of these cues, users rely on a range of semiotic resources, including typography, spelling variation, images, GIFs, emoticons, and emojis (see e.g. Kress and van Leeuwen 2006). These elements are not merely decorative. They rather play an essential role in shaping meaning by expressing stance, softening or intensifying messages, signaling identity, and framing how utterances should be interpreted (see also Yus 2011, 2025).

A critical insight of DPA is that although digital platforms provide globally uniform tools, their interpretation is anchored in local sociolinguistic contexts. In other words, the communicative value of semiotic resources depends heavily on cultural norms, localized dialectal practices, and community conventions (Yus 2025). As a result, what is perceived as polite, humorous, sarcastic, or offensive online varies significantly across social settings (Yu and Chang, 2024). Within this framework, this study treats emojis not as universal symbols with static semantics, but as context-sensitive pragmatic resources embedded within ADD. DPA provides the theoretical lens for examining how users adapt and assign meaning to emojis in alignment with local communicative expectations of social appropriateness, alignment, and identity performance (Shormani and Alenezi 2026).

## 2.3. Machine learning and digital pragmatics

An important conceptual intersection exists between computational modeling and digital pragmatics. Digital pragmatics provides the theoretical foundation for viewing online language as context-dependent, interactional, and functionally motivated, while ML offers computational procedures for detecting and modeling recurrent structural patterns across large collections of naturally occurring discourse (Jegede 2025; Shormani 2026). Their convergence goes beyond simply applying ML to linguistic datasets; it integrates pragmatic theory with automated pattern extraction. This synergy aligns with the broader paradigm of computational pragmatics, which applies formal computational methods to analyze relationships between linguistic expressions, communicative context, and social action (Jurafsky 2004; Yus 2025; Jegede 2025; Shormani 2026).

These 2 approaches fulfill complementary analytical objectives. Digital pragmatics addresses *what* a linguistic or multimodal feature signifies, *what* communicative function it fulfills, and *how* that function depends on interactional context. However, ML evaluates *whether* such patterns can be reliably detected, classified, or statistically modeled from observable linguistic data (Jurafsky 2004; Becker et al. 2020; Jegede 2025). In digital communication, this division of labor is crucial: online discourse produces massive amounts of multimodal data that are unfeasible to analyze manually at scale. Machine learning provides scalability and pattern detection, whereas digital pragmatics offers the interpretive framework necessary to understand those detected patterns as stance-taking, affect modulation, social alignment, or politeness strategies. This convergence has been demonstrated in computational studies in which pragmatically meaningful linguistic categories are operationalized through annotation and subsequently learned from textual patterns. For example, Becker et al. (2020) demonstrate how DL techniques can be used to classify heuristic textual practices in academic discourse, illustrating how theoretically informed pragmatic categories can be translated into computationally learnable representations. Such work is important for digital pragmatics because it demonstrates that pragmatic phenomena need not be restricted to close qualitative interpretation; they can also be operationalized, quantified, and examined across larger datasets. At the same time, computational classification does not by itself establish the pragmatic meaning of a pattern. Human linguistic and pragmatic interpretation remains necessary for determining what the detected patterns represent and why they may occur in particular communicative contexts (Becker et al. 2020).

The relationship can be extended further through the notion of the pragmatic web, which emphasizes the role of digital technologies and computational systems in shaping the production, interpretation, and circulation of meaning (Jones 2020). In digitally mediated environments, users interact not only with other users but also with algorithmically organized interfaces and systems that process behavioral and linguistic information. Consequently, digital pragmatic meaning may emerge from the interaction of human communicative choices, technological affordances, and algorithmic processing. From this perspective, ML is not merely an external analytical tool applied to digital discourse; it also forms part of the technological environment through which contemporary communication is increasingly processed and interpreted.

In our study, the purpose of integrating ML with DP is thus to connect computational learnability with pragmatic interpretation. Specifically, MARBERT is used to assess whether contextualized textual representations can reliably identify relevant lexical and pragmatic categories, while logistic regression provides interpretable estimates of the associations between these linguistic features and emoji occurrence. DP then provides the theoretical framework for interpreting these relationships as patterns of digitally mediated meaning-making rather than as purely statistical regularities (cf. Becker et al. 2020; Jones 2020). The intersection of the 2 approaches consequently enables the study to examine emoji use at complementary levels, viz., ML evaluates what can be computationally learned from the textual context, statistical modeling identifies measurable associations, and digital pragmatics explains how these associations may relate to communicative function and context. In this sense, the integration of ML and DP is not intended to replace pragmatic analysis, but to extend its empirical reach while retaining the interpretive principles necessary for understanding meaning in digital interaction (see also Jegede 2025).

### 2.4. Colloquial Arabic: Lexical vs pragmatic categories

CA represents a group of dialectal varieties used in everyday communication across the Arab world. Unlike SA, CA is primarily spoken and exhibits significant regional variation (see e.g. Holes 2004). Both SA and CA constitute what is so called diglossia (see e.g. Ferguson 1959; Shormani and Alenezi 2026). Versteegh (2014) highlights the diglossic nature of Arabic, where formal and informal varieties coexist with distinct functions. In digital communication, CA has gained prominence as users increasingly favor informal, expressive language. Bassiouney (2009) notes that CA allows for greater emotional and interpersonal expression compared to formal Arabic. This makes it particularly suitable for social media contexts, where personal voice and immediacy are valued. The use of CA in online platforms also reflects identity construction and social alignment. Sauter (2013) suggests that dialect choice can signal group membership and cultural affiliation. In this context, the interaction between CA and emojis may reveal important patterns of multimodal expression in Arabic digital discourse (ADD).

In this study, thus a distinction is made between linguistics/lexical features and discourse/pragmatic functions to better capture the multidimensional nature of emoji use in DC. Lexical categories including nouns, verbs, adjectives, and adverbs, primarily represent the propositional content of a message and are traditionally associated with clause-internal syntactic structure (cf. e.g. Chomsky 1957, 1981; Ouhalla 1999; Aoun et al. 2010; Shormani 2017, 2024). However, discourse/pragmatic categories such as interrogative and exclamative markers function at the interactional/informational structure or level of communication (cf. e.g. Rizzi 1997, 2004; Shormani 2017) where they encode speaker stance, affect, and communicative intent rather than propositional meaning. This distinction aligns with established work in corpus linguistics and pragmatics, which separates content words from discourse markers and interactional features that regulate interpersonal meaning (Biber et al. 1999; Aijmer 2013). Within digital discourse, such pragmatic elements have been shown to play a crucial role in expressing attitude and managing social interaction, particularly in online communication where multimodal resources such as emojis often co-occur with stance-taking devices (Kress and van Leeuwen 2006). Accordingly, this study treats question and exclamation forms as pragmatic indicators of interactional meaning, while lexical categories are interpreted as structural predictors of informational content. Thus, in this study, CA is conceptualized as any Arabic dialect acquired at home, i.e. other than SA, as has been noted so far (see also Versteegh 2014).

### 2.5. Emojis

Emojis are pictographic symbols used in DC to convey emotion, tone, and meaning to enhance interaction between interlocutors that may otherwise be difficult to express in text alone in DC (Yus 2025). Evans (2017) argues that emojis constitute a form of visual language that complements written text. They are not a replacement for language but an extension of it; they also compensate for face-to-face availability (Shormani and Alenezi 2026). Research has shown that emojis perform various communicative functions (Danesi 2016; Evans 2017), including emotional expression, emphasis, and discourse organization. Miller et al. (2016) demonstrate that emojis can alter the perceived meaning of text, highlighting their pragmatic significance. Moreover, emoji use is influenced by contextual and linguistic factors. It has been found that emojis are used systematically rather than randomly, often in relation to specific semantic fields (see e.g. Evans 2017; Yus 2025). This supports the premise that emoji distribution may be linked to lexical categories, a hypothesis central to this study.

Emojis have increasingly been conceptualized as integral components of digital communication rather than peripheral embellishments (Shormani 2026). Early work in DP highlights how digital communication lacks many of the non-verbal cues present in face-to-face interaction, leading users to adopt alternative strategies to convey affect and interpersonal meaning. Yus (2011) argues that users compensate for this absence through multimodal resources such as emojis, which function as pragmatic markers that guide interpretation and reduce ambiguity. Subsequent research has expanded this view by demonstrating that emojis play systematic roles in meaning construction. Evans (2017) proposes that emojis constitute a form of visual language that interacts with text to create enriched semantic and pragmatic effects. Similarly, Danesi (2016) emphasizes the semiotic nature of emojis, arguing that they operate as culturally shaped symbols that encode emotional and contextual meaning.

Thus, there is a gap this study attempts to address including the limitations of these studies by adopting a corpus-based approach to examine whether word category influences emoji use in colloquial ADD (cf. Li and Yang 2018; Wang et al. 2024). In doing so, it contributes to the growing field of digital pragmatics by providing empirical evidence of the interaction between textual and visual modes of communication.

## 3. Methods

### 3.1. Data collection: source and tools

The data for this study were collected from X, which is characterized by short-form, user-generated content and frequent use of multimodal elements such as emojis. This platform was selected due to its suitability for examining the interaction between textual and visual elements in naturally occurring digital discourse. Since most X users are free users, they are allowed to write posts of up to 280 characters per post (see X character counting documentation). The relatively short format was also considered useful for reducing potential ambiguity in classification, labeling, and tokenization. Data extraction was carried out using Python-based tools, including *tweepy* and *pandas*, which enabled the retrieval and processing of publicly available posts via API v2 in accordance with X policies. The collection process was conducted on September 18, 2025, without imposing a predefined temporal restriction, as the primary objective was to obtain a sufficiently large corpus of CA tweets. The search resulted in 28778 CA tweets, 15856 of which contained emojis.

### 3.2. Data refinement and normalization

After data collection, the dataset underwent a comprehensive refinement process to ensure quality and consistency. Text cleaning was performed using Python libraries such as *re* and *pandas*, including the removal of URLs, user mentions, identification parts in hashtags, and non-linguistic symbols such as &, *, %, and $. User names/account references, duplicate posts, empty posts, and irrelevant entries were also removed. Our aim of removing these elements was to ensure user privacy and to abide by X's conditions of use. As for normalization, Arabic orthographic variants were normalized, including *taa marboutah* ـةand the Arabic letter ـه. The purpose of this refinement was to improve the reliability and consistency of subsequent tokenization and linguistic classification. We have also refined the data from @useraccount, 3 dots …, and normalized Arabic variant elements including *taa marboutah* and *haa* into ـه. Data refinement was meant for enhancing the overall reliability of the datasets. Tokenization was meant to segment each post into individual lexical units, from which normalized features were derived: noun_norm, verb_norm, adj_norm, and adverb_norm as lexical features, and question_norm and exclamation_norm as pragmatic features. These normalized variables constitute the principal linguistic predictors used in the statistical (e.g. CAMEL POS) and predictive analyses (MARBERT standard metrics).

### 3.3. Pipeline

We adopt a supervised pipeline in this study as summarized in Fig 2. In the preprocessing stage, the collected tweets underwent several processes. These are detailed in the sections to follow:

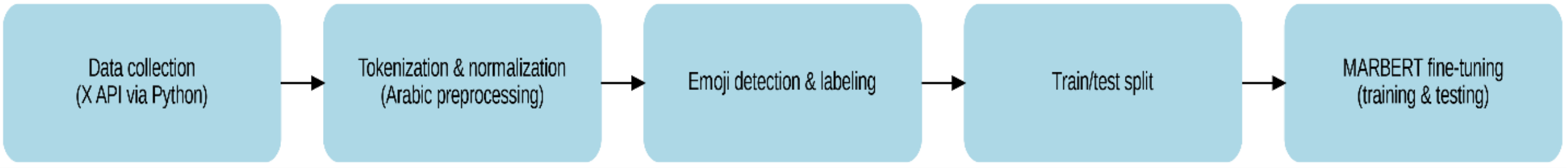

***Fig 2: Pipeline***

In the preprocessing stage, the collected texts underwent tokenization and normalization. They were tokenized into 4 lexical categories, i.e., noun_norm, verb_norm, adj_norm, adverb_norm, and 2 pragmatic categories, viz., question_norm, and exclamation_norm. After preprocessing, the dataset was then split into training and testing subsets to support supervised learning. During modeling, we install and configure all necessary libraries, CAMel, MARBERT in Google Colab, Python-based procedures were developed for preprocessing, modeling, and evaluation. The overall analytical procedure is represented in Fig 1.

File Edit Format Run Options Window Help

```
!pip install transformers torch camel-tools pandas scikit-learn -q

import pandas as pd
import numpy as np
import torch

from transformers import AutoTokenizer, AutoModelForSequenceClassification
from camel_tools.tokenizers.word import simple_word_tokenize
from camel_tools.pos import POSTagger

from sklearn.linear_model import LogisticRegression
from sklearn.model_selection import train_test_split
from sklearn.metrics import classification_report

df = pd.read_csv("CA_texts_emojis.csv")

model_name = "UBC-NLP/MARBERT"

tokenizer = AutoTokenizer.from_pretrained(model_name)
marbert_model = AutoModelForSequenceClassification.from_pretrained(model_name)

pos_tagger = POSTagger.pretrained()

def extract_pos_features(text):
    tokens = simple_word_tokenize(str(text))
```

*Fig 1: Part of Python script*

### 3.4. Data annotation and partitioning

Data annotation was conducted using the CAMeL tools toolkit, using POS tagging and morphological disambiguation capabilities. The procedure automatically processed the corpus at the token and sentence levels, assigning POS information to the linguistic units in each post and identifying relevant morphosyntactic categories, including nouns, verbs, adjectives, adverbs, interrogative forms, and exclamatory forms. The resulting annotations were then converted into feature variables: noun_norm, verb_norm, adj_norm, adverb_norm, question_norm, and exclamation_norm, representing the presence or frequency of these features in each post. This computational annotation provided a consistent and reproducible basis for examining the relationship between linguistic structure and emoji use, while substantially reducing the subjectivity and labor associated with manual annotation. Given that our emoji-data is large, a partitioning procedure, resulting in 3 non-overlapping datasets with 70% for training dataset, 15% for validation dataset, and 15% for evaluation dataset: 11099 tweets, 2378 tweets and 2378 tweets, respectively.

### 3.5. Model optimization and finetuning

MARBERT was optimized to ensure that the reported performance reflected genuine generalization rather than memorization of the training data. For the DP component, MARBERT was finetuned on the training dataset using the normalized textual representations and emoji-related information described above. The validation dataset was used during model development to monitor performance and guide optimization, whereas the evaluation dataset remained completely unseen until the final assessment. The model coefficients β indicate the direction and strength of each lexical category's effect on emoji use, where positive values suggest increased likelihood and negative values indicate decreased likelihood. (95%) Confidence Interval (CI) and

*p*-values were reported to assess the statistical reliability and significance of the effects. For D component, the training-set parameters were used exclusively for feature standardization. MARBERT classification performance was evaluated using *accuracy, precision, recall, F1-score,* and *weighted F1-score*. These measures were calculated from the numbers of true positives (TP), true negatives (TN), false positives (FP), and false negatives (FN) obtained for each linguistic category. Accuracy represents the proportion of correctly classified instances among all evaluated instances:

$$Accuracy = \frac{TP+TN}{TP+TN+FP+FN}$$

Precision measures the proportion of instances predicted as a given category that are correctly classified:

$$Precision = \frac{TP}{TP+FP}$$

Recall measures the proportion of actual instances of a category that are correctly identified by the model:

$$\text{Recall} = \frac{TP}{TP+FN}$$

F1-score provides a harmonic mean of precision and recall and is particularly useful when the distribution of categories is imbalanced:

$$\text{F1-score } F1 = 2 \times \frac{Precision \times Recall}{Precision + Recall}$$

For the multi-category evaluation, the weighted F1-score was calculated by weighting the F1-score of each category according to its support, where support represents the number of true instances belonging to that category:

$$\text{Weighted-F1} = \sum_{k=1}^{K} \frac{n_k}{N} F1_k$$

where $K$ is the total number of categories, $n_k$ is the number of true instances (i.e. support) for category $k$, $F1_k$ is the F1-score for category $k$, and $N$ is the total number of evaluated instances. These metrics provide complementary perspectives on model performance. *Accuracy* reflects overall classification correctness, whereas *precision* and *recall* distinguish between the reliability and completeness of category identification. F1 balances these 2 dimensions, while weighted-F1 accounts for the observed distribution of categories by assigning greater weight to categories with larger support. Given the substantial imbalance among the linguistic categories in ‘this dataset, reporting per-category performance together with weighted-F1 provides a more informative evaluation than relying on accuracy alone.

**4. Results**

In this section, we analyze the study results. These results are presented in Tables 1-3, for descriptive statistics, Logistic regression and coefficient, and MARMERT standard performance.

## Table 1: Descriptive statistics

| Term | **noun_norm** | **verb_norm** | **adj_norm** | **adv_norm** | **question_norm** | **exclamation_norm** |
|---|---|---|---|---|---|---|
| **mean** | 0.6749 | 0.0831 | 0.0011 | 0.0013 | 0.0026 | 0.0111 |
| **std** | 0.1606 | 0.0997 | 0.0103 | 0.0005 | 0.0176 | 0.0348 |
| **min** | 0.0000 | 0.0000 | 0.0000 | 0.0000 | 0.0000 | 0.0000 |
| **25%** | 0.5789 | 0.0000 | 0.0000 | 0.0000 | 0.0000 | 0.0000 |
| **50%** | 0.6667 | 0.0588 | 0.0000 | 0.0000 | 0.0000 | 0.0000 |
| **75%** | 0.7778 | 0.1277 | 0.0000 | 0.0000 | 0.0000 | 0.0000 |
| **max** | 1.0000 | 1.0000 | 0.2500 | 0.1106 | 0.5000 | 0.3333 |

## Table 2: Logistic regression and coefficient

| **Variable** | **Coef (β)** | **Std. Er** | **z-score** | **p-value (P>\|z\|)** | **Cl_L** | **Cl_U** |
|---|---|---|---|---|---|---|
| **const** | 4.6325 | 0.4230 | 10.952 | 0.001 | 3.803 | 5.462 |
| **noun_norm** | -1.4774 | 0.4481 | -3.297 | 0.001 | -2.356 | -0.599 |
| **verb_norm** | 0.8206 | 0.2494 | 3.290 | 0.001 | 0.332 | 1.309 |
| **adj_norm** | 0.3592 | 0.1000 | 3.592 | 0.001 | 0.163 | 0.555 |
| **adverb_norm** | -1.3029 | 0.3500 | -3.723 | 0.001 | -1.989 | -0.617 |
| **question_norm** | 0.5044 | 0.1500 | 3.363 | 0.001 | 0.210 | 0.798 |
| **exclamation_norm** | 0.5044 | 0.5704 | 0.884 | 0.377 | -0.614 | 1.622 |

## Table 3: MARMERT per-class performance

| **Category** | **Accuracy** | **Precision** | **Recall** | **F1** | **Weighted F1** |
|---|---|---|---|---|---|
| noun_norm | 0.58 | 0.88 | 0.44 | 0.59 | 0.58 |
| verb_norm | 0.76 | 0.74 | 0.99 | 0.85 | 0.71 |
| adj_norm | 0.33 | 0.00 | 0.00 | 0.00 | 0.16 |
| adverb_norm | 0.34 | 0.76 | 0.02 | 0.05 | 0.19 |
| question_norm | 0.33 | 0.00 | 0.00 | 0.00 | 0.16 |
| exclamation_norm | 0.33 | 0.00 | 0.00 | 0.00 | 0.16 |

As shown in Table 1, nouns overwhelmingly dominate the corpus, with the highest mean normalized frequency (M = 0.675, SD = 0.161), followed by verbs (M = 0.083, SD = 0.100). Adjectives (M = 0.001, SD = 0.010), adverbs (M = 0.001, SD = 0.001), questions (M = 0.003, SD = 0.018), and exclamations (M = 0.011, SD = 0.035) occur substantially less frequently. The median is zero for adjectives, adverbs, questions, and exclamations, whereas nouns and verbs have median values of 0.667 and 0.059, respectively.

Table 2 shows significant negative associations between emoji occurrence and noun density (β = −1.477, p = .001, 95% CI [−2.356, −0.599]) and adverb density (β = −1.303, p = .001, 95% CI [−1.989, −0.617]). However, verb density (β = 0.821, p = .001, 95% CI [0.332, 1.309]), adjective density (β = 0.359, p = .001, 95% CI [0.163, 0.555]), and question density (β = 0.504, p = .001, 95% CI [0.210, 0.798]) show significant positive associations with emoji occurrence. Exclamation density is positive but statistically non-significant (β = 0.504, p = .377, 95% CI [−0.614, 1.622]). These findings demonstrate that frequency and statistical association with emoji occurrence do not necessarily coincide.

Table 3 provides MARBERT's category-level classification performance. The model performs best for verbs (precision = .74, recall = .99, F1 = .85), followed by nouns (precision = .88, recall = .44, F1 = .59). Adverbs achieve high precision (.76) but very low recall (.02), yielding an F1 score of .05, whereas adjectives, questions, and exclamations receive zero precision, recall, and F1. Overall, the findings show that nouns are the most frequent category, while verbs demonstrate the strongest MARBERT classification performance and a significant positive association with emoji occurrence. The sparse categories remain substantially more difficult for the model to identify reliably.

## 5. Discussion

Given our scope and purpose, and based on our findings in the above section, the study seems to bridge ML and digital pragmatics by showing how computational methods can reveal systematic patterns in language use, but also highlight their limitations in capturing the complexity of human communication on ADD. Emoji use emerges as a multimodal phenomenon shaped by linguistic structure and discourse function, requiring approaches that combine statistical modeling with theoretical insights from linguistics and communication studies (cf. Li and Yang 2018; Ullah et al. 2025). This framework combines binary logistic regression and MARBERT finetuning. Logistic regression provides interpretable evidence about the statistical associations between predefined linguistic categories and emoji occurrence, whereas MARBERT evaluates the extent to which these categories can be identified from contextual textual representations. The 2 approaches thus address related but distinct dimensions of emoji-related language use.

The regression results reveal clear differences across linguistic categories. Noun density is significantly negatively associated with emoji occurrence ($\beta = -1.477$, $p = .001$; 95% CI [−2.356, −0.599]), whereas verb density is significantly positively associated with emoji occurrence ($\beta = 0.821$, $p = .001$; 95% CI [0.332, 1.309]). This contrast suggests that noun-dense discourse, which may be more referential or information-oriented, is less likely to co-occur with emojis, whereas verb-dense discourse is more likely to do so. The remaining predictors provide further evidence that emoji occurrence is not determined by lexical frequency alone. Adjective density is significantly positively associated with emoji occurrence ($\beta = 0.359$, $p = .001$; 95% CI [0.163, 0.555]), whereas adverb density is significantly negatively associated ($\beta = -1.303$, $p = .001$; 95% CI [−1.989, −0.617]). Question density also shows a significant positive association ($\beta = 0.504$, $p = .001$; 95% CI [0.210, 0.798]). However, exclamation density has a positive but statistically non-significant association ($\beta = 0.504$, $p = .377$; 95% CI [−0.614, 1.622]). The latter result cautions against interpreting exclamatory structures as reliable predictors of emoji occurrence in 'this dataset. Collectively, these findings suggest that emoji occurrence is associated with the structural and functional characteristics of the surrounding discourse rather than with the frequency of individual linguistic categories alone.

The MARBERT results complement the regression findings but should not be interpreted as direct evidence for the regression associations. MARBERT performs particularly well in identifying verbs (accuracy = .76, precision = .74, recall = .99, F1 = .85) and nouns (accuracy = .58, precision = .88, recall = .44, F1 = .59), while performance is substantially weaker for the less frequent categories. Adverbs achieve relatively high precision (.76) but extremely low recall (.02), resulting in an F1 of .05. Adjectives, questions, and exclamations receive zero precision, recall, and F1. This disparity highlights an important methodological distinction: a statistically significant association between a linguistic feature and emoji occurrence does not necessarily imply that MARBERT can reliably identify that category from textual input. Thus, regression significance and ML classification performance represent different dimensions of evidence.

The predominance of nouns in the dataset is also consistent with observations concerning the prevalence of nominal and SVO structures in Arabic varieties (cf. Mohammad 2000; Holes 2004). This may also have to do with the nature of CA and Arabic in general that MARBERT has been trained on. Verbless/nominal sentences in present tense contain null verbs (which are actually verbs but not spelled out, see e.g. Al-Balushi 2012). These sentences are also referred to as topic-comment sentences or المبتدأ والخير sentences (see e.g. Shormani 2017), the verb does not show up in imperfective mode, but it does in perfective one as in الطالبُ مجتهدٌ *at-taalibu mujtahidun* ‘the student is hardworking’ where the verb, precisely *be* is null, while in perfective mode كان الطالبُ مجتهداً, *kaan at-taalibu mujtahidan* the verb shows up as *be* كان. This argument is actually discourse-based, which is based on a discourse maxim, given Grice’s Maxim of Quantity (GMQ). GMQ states that an interlocuter cannot say more or less than it is required. Thus, given GMQ it may well be argued that MARBERT may have considered null verbs in CA. This could be taken to indicate that frequency should not be equated with statistical influence: nouns are by far the most frequent category, yet their association with emoji occurrence is negative, whereas the less frequent verbs show a positive association (cf. also Chomsky 1981).

These findings also demonstrate the value of combining interpretable statistical modeling with contextual representation. MARBERT, as a transformer-based architecture developed for Arabic dialectal and social-media text, can capture contextual and semantic information extending beyond manually specified linguistic features. Its relatively strong performance for verbs suggests that contextual representations can identify this category reliably in ‘this corpus. However, the weak performance for adjectives, adverbs, questions, and exclamations indicates that contextual modeling remains sensitive to category frequency and representation. It then follows that MARBERT should not be interpreted as establishing which linguistic categories statistically influence emoji occurrence. It rather provides complementary evidence concerning the model’s ability to recognize those categories from contextualized text.

From a DPA, the findings suggest that emojis are not randomly distributed across ADD. Their occurrence is associated with particular linguistic configurations, although the direction and reliability of these associations vary across categories. The significant negative association of noun density may be compatible with a distinction between more referential or information-oriented discourse and more expressive or interactional communication. Conversely, the significant positive association of verb density may reflect the compatibility of action-oriented language with contextual and interactional meaning-making. The positive associations observed for adjective and question densities further suggest that emoji occurrence can accompany linguistic structures involved in description, evaluation, and interaction. These interpretations should, however, be understood as associational rather than causal; ‘this analysis does not establish that any linguistic category causes emoji use.

The integration of MARBERT and logistic regression therefore provides 2 complementary perspectives on emoji-related language use. Logistic regression identifies statistically interpretable associations between linguistic features and emoji occurrence, whereas MARBERT evaluates how reliably the corresponding linguistic and pragmatic categories can be identified from contextualized text. The contrast between the 2 sets of results (Tables 2 & 3) is particularly informative, i.e. verbs are significantly positively associated with emoji occurrence and are also the category most reliably identified by MARBERT (F1 = .85), whereas some statistically significant predictors, particularly adjectives and questions, remain difficult for the model to identify reliably. This distinction demonstrates that linguistic frequency, statistical association, and computational predictability are not equivalent measures and should therefore not be interpreted interchangeably.

From a DPA perspective, the study conceptualizes CA as a context-sensitive form of meaning-making in which linguistic choices, multimodal resources such as emojis, and interactional purposes are interconnected. Within this framework, emojis are treated not simply as decorative elements but as pragmatic resources that can contribute to stance-taking, emotional expression, and discourse management in digital communication. The observed statistical patterns consequently provide evidence that different linguistic configurations are associated with different probabilities of emoji occurrence. In particular, the significant negative association between noun density and emoji occurrence is compatible with a greater orientation toward referential or information-focused discourse, whereas the significant positive association between verb density and emoji occurrence may reflect the role of dynamic and action-oriented language in expressive and interactional communication. The significant positive associations of adjective and question densities likewise indicate that emojis can occur alongside descriptive, evaluative, and interactional structures. Exclamation density, by contrast, shows a positive but non-significant association and therefore does not provide sufficient statistical evidence for a reliable effect in 'this dataset.

The integration of MARBERT and logistic regression thus provides complementary evidence at two analytical levels. MARBERT captures contextual patterns in the textual input and evaluates the model's ability to identify the relevant categories, while logistic regression provides interpretable estimates of the associations between normalized linguistic/pragmatic features and emoji occurrence. The findings support the view that emoji use in Arabic X discourse is systematically related to linguistic structure and contextualized communication, rather than being adequately explained by lexical frequency alone. In this sense, the hybrid approach contributes to digital pragmatics by connecting computational prediction with interpretable linguistic evidence. At the same time, the contrasting classification results demonstrate that statistical significance does not necessarily translate into reliable ML prediction, particularly for sparsely represented categories.

To recapitulate, nouns constitute the dominant lexical category in the dataset, yet noun density is significantly negatively associated with emoji occurrence. Verbs, although substantially less frequent, show a significant positive association and the strongest MARBERT classification performance (F1 = .85). Adjective and question densities also show significant positive associations, whereas adverb density is significantly negative. Exclamation density is positive but non-significant. These findings indicate that emoji deployment in ADD is not a random stylistic flourish, but a systematically patterned phenomenon deeply tethered to the morphosyntactic and pragmatic architecture of the sentence. This is supported by both computational and statistical measures. Put simply, combining interpretable binary logistic regression with MARBERT finetuning, the study highlights that statistical association and ML standardization represent distinct dimensions of digital communication, proving that emojis function as vital pragmatic resources for stance-taking and interpersonal, affective and interactional coordination rather than mere decorative appendages (Yus 2025; Shormani 2026).

## 6. Conclusions and limitations

To conclude, this study examines ML in relation to digital pragmatics and investigates whether linguistic categories such as nouns, verbs, adjectives, adverbs, questions, and exclamations systematically influence emoji use. It employs MARBERT and logistic regression within a digital-pragmatics framework. 15856 tweet-emoji texts constituted the main emoji-focused dataset, and were used for both models: logistic regression and coefficient, and MARBERT analyses. Findings reveal that emoji occurrence is systematically related to linguistic structure and discourse function rather than being randomly distributed across textual contexts. Thus, a number of conclusions can be drawn from this study including: i) emoji use in ADD is most effectively understood through a hybrid analytical framework that integrates MARBERT-based contextual learning with interpretable lexical-statistical modeling, ii) while MARBERT captures contextual and semantic patterns underlying linguistic-category identification, logistic regression provides transparent evidence of how specific linguistic structures are systematically associated with emoji occurrence, iii) emoji use is not randomly distributed across linguistic contexts but is shaped by word-category distributions and discourse/pragmatic structures, iv) the significant negative relationship between noun density and emoji use indicates that referential and information-oriented discourse is less likely to be accompanied by emojis, whereas the significant positive relationship between verb density and emoji use suggests that action-oriented and dynamic discourse is more compatible with expressive multimodal communication, and v) lexical categories seem to be accompanied by emojis more than pragmatic categories do .

The findings further show MARBERT performs comparatively well for verbs (precision = .74, recall = .99, F1 = .85) and nouns (precision = .88, recall = .44, F1 = .59), whereas performance is substantially weaker for the less frequent categories. Adverbs, for example, achieve high precision but very low recall and F1 (precision = .76, recall = .02, F1 = .05), which could be linked to their linguistic nature, as adverbs describe the manner of performing actions by verbs (Yule 2014), while adjectives, questions, and exclamations receive zero precision, recall, and F1. This distinction demonstrates that statistical significance does not necessarily translate into reliable ML prediction, particularly for sparsely represented linguistic categories. From a digital-pragmatic perspective, these findings support the view that emojis function as pragmatic and affective markers rather than merely stylistic or ornamental elements. Their distribution reflects underlying communicative strategies in Arabic X discourse, particularly the interaction between referential, informational, expressive, and interactional forms of communication. Emojis therefore operate as multimodal cues that contribute to meaning, stance, emotional expression, emphasis, and interactional alignment in online discourse.

However, there are several limitations involved in this study including: i) reliance on automatic linguistic annotation may introduce potential inaccuracies, particularly given the morphological complexity of Arabic and the informal, variable nature of social media language, ii) the dataset is restricted to X, which may constrain the generalizability of the findings to other digital platforms characterized by different communicative norms, user practices, and technological affordances, iii) the substantial differences in MARBERT performance across categories indicate that sparse linguistic features remain difficult to identify reliably, even when their statistical associations with emoji occurrence are significant, and iv) the relatively heterogeneous effects across predictors demonstrate that lexical features alone cannot fully capture the contextual and interactional complexity of emoji use.

Based on the limitations, we propose that future studies could incorporate richer contextual representations, discourse-level modeling, and interactional metadata, including reply structure, user engagement, conversational position, and temporal dynamics. Although machine annotation avoids subjectivity, human annotation could also be involved to avoid some inconsistencies. Cross-platform comparative studies could also be valuable for examining how interface design and communicative affordances shape emoji behavior across different digital environments. Further research could additionally explore larger and more balanced datasets to improve the representation of sparse linguistic categories and enhance the reliability of ML classification. From a ML perspective, the findings suggest that interpretable feature-based models remain valuable for identifying and explaining systematic linguistic associations, but they should be complemented by deep contextual models capable of capturing broader semantic, pragmatic, and discourse-level patterns.